\documentclass{IOS-Book-Article}

\usepackage{nesybook}

\usepackage{chapter-dummy/chap}

\begin{document}

\begin{frontmatter}              

\title{Compendium of Neuro-Symbolic Artificial Intelligence}


\author[A]{\fnms{Pascal} \snm{Hitzler}}, 
\author[B]{\fnms{Md Kamruzzaman} \snm{Sarker}},
and 
\author[A]{\fnms{Aaron} \snm{Eberhart}}

\runningauthor{Pascal Hitzler, Md Kamruzzaman Sarker, Aaron Eberhart}
\address[A]{Kansas State University}
\address[B]{Kansas State University}

\end{frontmatter}

\maketitle

\setcounter{tocdepth}{0}
\tableofcontents

\setcounter{page}{0}

\chapter{Neuro-Symbolic Spatio-Temporal Reasoning}
\label{cha:neuro-symb-spat}

\chapterauthor{Jae Hee Lee\astfootnote{\label{contrib}Equal contribution}}{University of Hamburg}
\chapterauthor{Michael Sioutis\textsuperscript{*}}{LIRMM UMR 5506, University of Montpellier, CNRS}
\chapterauthor{Kyra Ahrens}{University of Hamburg}
\chapterauthor{Marjan Alirezaie}{Örebro University}
\chapterauthor{Matthias Kerzel}{University of Hamburg}
\chapterauthor{Stefan Wermter}{University of Hamburg}

\allchapterauthors{Jae Hee Lee, Michael Sioutis, Kyra Ahrens, Marjan Alirezaie, Matthias Kerzel, Stefan Wermter}

\section{Introduction}
\label{sec:intro}

Knowledge about space and time is an essential prerequisite for solving problems in the physical world: An artificial intelligence (AI) agent, when situated in a physical environment and interacting with objects, often needs to reason about positions of and relations between objects; and as soon as the agent plans its actions to solve a task, it needs to consider the temporal aspect (e.g., what actions to perform over time). Spatio-temporal knowledge, however, is required beyond interacting with the physical world, and is also often transferred to the abstract world of concepts through analogies and metaphors (e.g., ``a threat that is hanging \emph{over} our heads''). As spatial and temporal reasoning is ubiquitous, different attempts have been made to integrate this into AI systems.

In the area of knowledge representation, spatial and temporal reasoning has been largely limited to modeling objects and relations and developing reasoning methods to verify statements about them~\cite{dylla_survey_2017,ligozat2011qualitative,cohn_qualitative_2008,DBLP:reference/spatial/2007}; even so, it has found its way into applications in many areas and domains that include 
cognitive robotics~\cite{DBLP:journals/jirs/DyllaW07}, 
qualitative model generation from video~\cite{DBLP:journals/jair/DubbaCHBD15}, ambient intelligence~\cite{DBLP:conf/cosit/BhattDH09}, 
and data mining~\cite{DBLP:journals/kais/MoskovitchS15a}. 
 On the other hand, neural network researchers have tried to teach models to learn spatial relations from data despite limited reasoning capabilities~\cite{hong_ptr_2021,goyal_rel3d_2020,johnson_clevr_2017}. Bridging the gap between these two approaches in a mutually beneficial way could allow us to tackle many complex tasks, such as natural language processing, visual question answering, and semantic image segmentation~(cf.~\cite{wermter_integration_1989,lee_what_2022,DBLP:journals/semweb/AlirezaieLSL19}).

 In this chapter, we examine the aforementioned integration problem from the perspective of neuro-symbolic AI. Specifically, we propose a synergy between logical reasoning and machine learning (ML) that will be grounded on spatial and temporal knowledge. This chapter is organized as follows: In Section~\ref{sec:background}, we provide a brief overview of neuro-symbolic AI and qualitative spatio-temporal reasoning. Then, in Section~\ref{sec:neuro-symb-spat}, we introduce some works on neuro-symbolic spatio-temporal reasoning and learning, namely \emph{a semantic referee for geospatial semantic segmentation of satellite imagery data} (Section~\ref{sec:relwork}) and \emph{a symbolic meta-learning interface for learning relative directions} (Section~\ref{sec:symb-meta-learn}), and propose a generic neuro-symbolic framework for integrating qualitative spatio-temporal reasoning and neural methods from a probabilistic perspective. Within that section, we also review recent datasets for spatial relation learning. Finally, in Section~\ref{sec:conclusion} we conclude this chapter and provide new directions for future research.


\section{Background}
\label{sec:background}

\subsection{Neuro-Symbolic AI}
\label{sec:nesy}

\begin{figure}[t]
    \centering \includegraphics[width=0.9\textwidth]{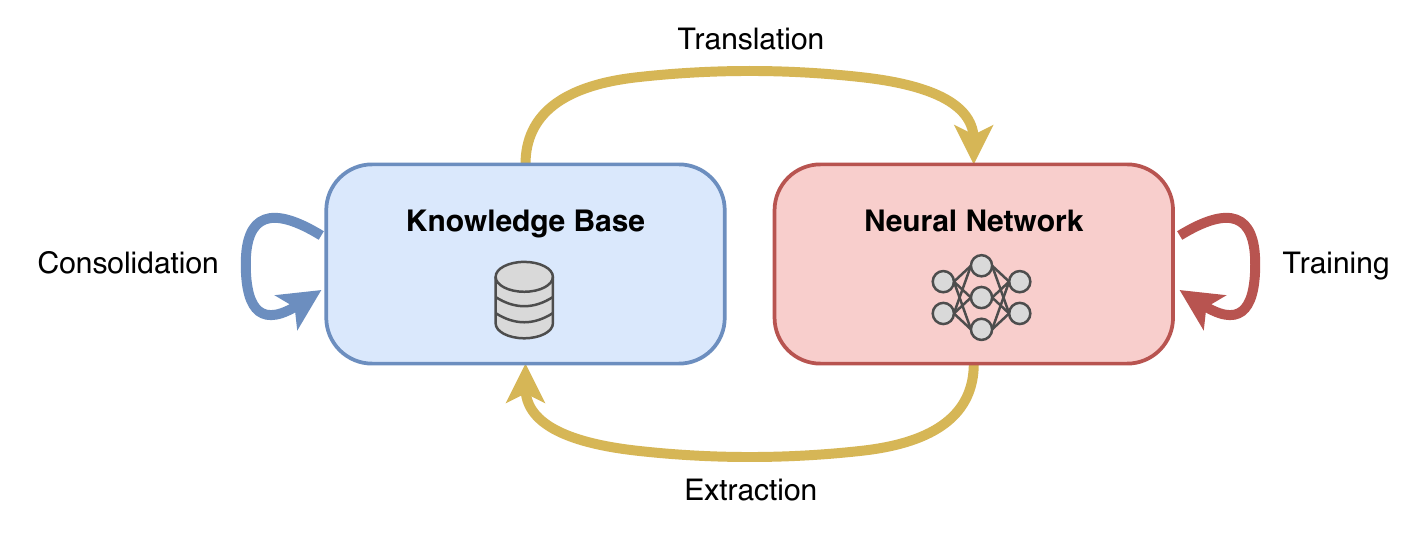}
\caption{\label{fig:nesycycle} Cyclical interaction in neuro-symbolic AI; a symbolic
system feeds symbolic (partial) knowledge to a neural network
system, which can be trained on raw data, and knowledge acquired through machine learning can then be extracted and fed back to the symbolic system, and made available for further processing in symbolic form (cf.~\cite[Figure~$1$]{DBLP:conf/birthday/BaderH05}).}
\end{figure}

One of the main challenges in AI today, and where the most progress towards the AI dream is expected to be seen over the next decade, is the seamless integration of statistical learning and symbolic reasoning~\cite{DBLP:conf/fsttcs/Valiant08,michael_l_littman_gathering_2021}. Such an integration is the area of study of neuro-symbolic AI~\cite{hitzler_neuro-symbolic_2021,326,wermter_overview_2000,mcgarry_hybrid_1999}.

Specifically, over the past years, statistical learning and symbolic reasoning have been developed mostly separately in AI. 
What is more, the representations (models) that are learned in statistical learning are generally low-level (sub-symbolic), whereas the representations used in symbolic reasoning are high-level (symbolic); thus, integrating the former with the latter representations is challenging, and it is exactly this task that neuro-symbolic AI addresses. To sum up, neuro-symbolic AI seeks to combine principles from neural-network learning and logical reasoning by leveraging the strengths of both worlds to the extent possible. In a sense, this framework closely resembles how humans perform problem-solving, as, from a psychological viewpoint, the perception, which is a data-driven process, and the logical reasoning, which is a knowledge-driven process, are entangled rather than separated in humans~\cite{DBLP:journals/chinaf/Zhou19}.

In general, we can classify neuro-symbolic approaches as follows~\cite[Figure~$24$]{DBLP:journals/ai/BadreddineGSS22}:
\begin{itemize}

    \item \emph{Neural architectures for logical reasoning}, e.g.,~\cite{DBLP:journals/jair/CohenYM20} where each clause of a stochastic logic program is converted into a factor graph with reasoning becoming differentiable so that it can be implemented by deep networks;

    \item \emph{Logical specification of neural networks}, e.g.,~\cite{DBLP:journals/jair/SourekAZSK18} where a Datalog program is used as a compact specification of a diverse range of existing advanced neural architectures, with a particular focus on Graph Neural Networks (GNNs) and their generalizations;

    \item \emph{Architectures integrating learning and reasoning}, e.g.,~\cite{DBLP:journals/ai/ManhaeveDKDR21} where a neural probabilistic logic programming language is introduced that incorporates deep learning by means of neural predicates.

\end{itemize}

A comprehensive review of the aforementioned classes of neuro-symbolic AI approaches can be found for example in~\cite{DBLP:journals/ai/BadreddineGSS22,hitzler_neuro-symbolic_2022,sarker_neuro-symbolic_2021,hitzler_neuro-symbolic_2021}. Here, we focus on the third of these classes, i.e., on \emph{architectures integrating learning and reasoning}, a high-level illustration of which is shown in Figure~\ref{fig:nesycycle}, along with all aspects pertaining to this class, such as evaluation datasets, tools, challenges, and so on. In particular, we further extend the perspective from architecture integration to the integration into the whole training procedure, which includes the aspect of how the datasets are prepared (e.g., dataset augmentation, multi-task learning) and how the examples from a dataset are presented to a model (e.g., curriculum learning) that help improve the performance of a model. In conclusion, in this chapter, we address this more generic type of neuro-symbolic approaches (cf. Section~\ref{sec:symb-meta-learn}).

\subsection{Qualitative Spatio-Temporal Reasoning}
\label{sec:qstr}

\begin{figure}[t!]
\centering
\scriptsize
{}\hfill
         \begin{tikzpicture}[scale=0.7]
         \begin{scope}[yshift=4mm,xshift=5mm,scale=0.7]
         \draw[rounded corners] (0, 1.2) rectangle (4, 2.9) {};
         \draw[thin] (2,2) circle (0.55);
         \draw (2,2) circle (0.7);
         \draw[ultra thick] (1,1.35) -- +(0,0.9);
         \draw[ultra thick] (0.8,2.25) -- +(0.4,0.0);
         \foreach \x in {0,1,2} {
         \draw[thick] (0.80+\x*0.2,2.25) -- + (0,0.4);
         }
         \draw[ultra thick] (3,1.35) -- +(0,1.3);
         \draw[thick] (3,2.65) -- (2.8,2.5) -- (2.8, 2.0) -- (3, 1.8);
         \end{scope}
         
         \node[draw,ellipse] (f) at (0,0) {fork};
         \node[draw,ellipse] (p) at (2,0) {plate};
         \node[draw,ellipse] (k) at (4,0) {knife};
         \node[draw,ellipse] (t) at (2,-1.5) {table};
         \draw[->] (f) to [bend right=45]  node [midway,fill=white] {\emph{on}} (t);
         \draw[->] (k) to [bend left=45]  node [midway,fill=white] {\emph{on}} (t);
         \draw[->] (p) -- (t)  node [midway,right] {\emph{on}} (t);
         \draw[->] (p.north) to [bend right=20]  node [midway,above,yshift=-1mm] {\emph{right of}} (f.north);
         \draw[->] (p.north) to [bend left=20]  node [midway,above,yshift=-1mm] {\emph{left of}} (k.north);
         \end{tikzpicture}
\hfill%
        \begin{tikzpicture}[scale=.5]
          \node[draw,ellipse] (x1) at (0,0) {Task~$A$};
          \node[draw,ellipse] (x2) at (8,0) {Task~$C$};
          \node[draw,ellipse] (x3) at (4,4) {Task~$B$};
          \draw [dotted,-] (x1) -- (x2) node [midway, fill=white] {$?$};
         \draw [->] (x1) -- (x3) node [midway, fill=white] {\emph{precedes}  $\vee$ \emph{follows} };
          \draw [->] (x3) -- (x2) node [midway, fill=white] {\emph{precedes}};
         \end{tikzpicture}
         ~~~~~~~~
\caption{Left: Qualitative abstraction of a spatial configuration. Right: A simplified temporal constraint network of $3$ variables, each representing a task; `$?$' denotes complete uncertainty\label{fig:simpleqcn}}.
\end{figure}
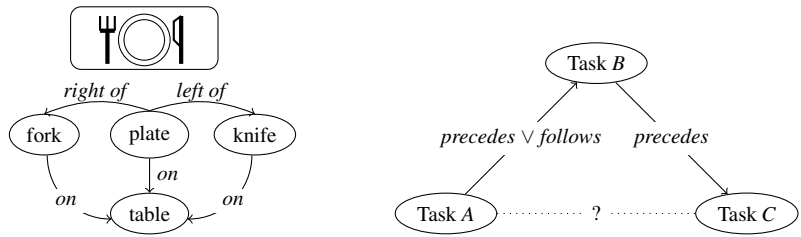

In everyday natural language descriptions, one typically uses expressions such as
\emph{outside}, \emph{left of}, or \emph{during} to spatially or temporally relate one object with another
object or oneself, without necessarily providing the exact metric information about these entities. An AI 
framework that aims to capture this type of human-like representation and reasoning pertaining to space and time is known in the research community
as Qualitative Spatial and Temporal Reasoning (QSTR)~\cite{dylla_survey_2017,ligozat2011qualitative}.
Specifically, QSTR is a major field of study in Symbolic
AI, and in particular in Knowledge Representation and Reasoning, that deals with the fundamental
cognitive concepts of space and time in an abstract, natural manner, ranging from theoretical computer science and logic to practical algorithms and
applications~\cite{dylla_survey_2017,ligozat2011qualitative}. 
More formally, QSTR restricts the vocabulary of rich mathematical theories that deal
with spatial and temporal entities to simple qualitative constraint languages, which can be used to form
interpretable spatio-temporal constraint networks of disjunctions of 
base relations, such as the one shown in Figure~\ref{fig:simpleqcn}. QSTR forms a concise framework that allows for rather inexpensive reasoning about entities located in space and time, and hence further boosts
research and applications in areas such as 
cognitive robotics~\cite{DBLP:journals/jirs/DyllaW07}, 
qualitative model generation from video~\cite{DBLP:journals/jair/DubbaCHBD15}, ambient intelligence~\cite{DBLP:conf/cosit/BhattDH09}, visual 
sensemaking~\cite{DBLP:conf/ijcai/SuchanBV19}, 
data mining~\cite{DBLP:journals/kais/MoskovitchS15a,DBLP:journals/datamine/KostakisTG17}, 
and qualitative case-based reasoning and learning~\cite{DBLP:journals/ai/HomemSCBM20}. 

To facilitate discussion, we first recall the formal definition of a \emph{qualitative constraint language}, which is a constraint language that is used to represent and reason about qualitative information.
A binary qualitative spatial or temporal constraint language is based on a finite set
$\B$ of \emph{jointly exhaustive and pairwise disjoint} relations, called \emph{base relations}~\cite{ligozat2011qualitative} and defined over an infinite domain $\D$.
The base relations of a particular qualitative constraint language can be used to represent the definite
knowledge between any two of its entities with respect to the level of granularity provided by the domain $\D$. The set $\B$ contains the identity relation ${\mathsf{Id}}$,
and is closed under the \emph{converse} operation ($^{-1}$). Indefinite knowledge can
be specified by a union of possible base relations, and is represented by the set containing them.
Hence, $2^{\B}$ represents the total set of relations. The set $2^{\B}$ is equipped with the usual set-theoretic operations of union and intersection,
the converse operation, and the \emph{weak composition} operation denoted by the symbol $\diamond$~\cite{ligozat2011qualitative}.
For all $r \in 2^\B$, we have that $r^{-1} = \bigcup\{b^{-1}~|~b \in r\}$.
The weak composition $(\diamond)$ of two base relations $b,b^\prime \in \B$
is defined as the smallest (i.e., most restrictive) relation $r \in 2^{\B}$ that includes $b \circ b^\prime$, or, formally,
$b \diamond b^\prime {=} \{ b^{\prime\prime} \in \B~|~b^{\prime\prime}{\cap}(b \circ b^\prime) \neq \emptyset \}$, where
$b \circ b^\prime {=} \{ (x, y) \in \D\times{\D}~|~\exists{z \in \D}~\text{such that}~
( x, z) \in b \wedge (z, y) \in b^\prime \}$ is the (true) composition of $b$ and $b^\prime$.
For all $r,r^\prime \in 2^\B$, we have that $r \diamond r^\prime$ $=$  $\bigcup\{b\diamond{b^\prime}~|~b\in{r},~b^\prime\in{r^\prime}\}$. We note that the weak composition
operation is of particular importance as it is the core operation used to perform some basic inference/reasoning, e.g., $\text{inside}(x,z)$ $\diamond$ $\text{inside}(z,y)$ $\to$ $\text{inside}(x,y)$,
and consequently check the admissibility/validity of large complex networks defined over disjunctions of base relations~\cite{IJCAI2016,sioutis2018} (see also Definition~\ref{def:qcn} later on).

\begin{figure}[t!]
\scriptsize
\centering
\begin{tabular}{ccc}
precedes $p$ & meets $m$ & overlaps $o$\tabularnewline
\begin{tikzpicture}
\draw[<->] (0,0) -- node[below] {$x$} (1,0);
 \draw[<->] (1.5,0) -- node[below] {$y$} (2.5,0);
 \end{tikzpicture} &
 \begin{tikzpicture}
\draw[<->] (0,0) -- node[below] {$x$} (1,0);
 \draw[<->] (1,0) -- node[below] {$y$} (2,0);
 \end{tikzpicture} &
 \begin{tikzpicture}
 \draw[<->] (0,0) -- node[below] {$x$} (1.5,0);
 \draw[<->] (1,0) -- node[below] {$y$} (2.5,0);
 \end{tikzpicture}\tabularnewline
preceded by $pi$ & met by $mi$ & overlapped by $oi$\tabularnewline
\begin{tikzpicture}
\draw[<->] (0,0) -- node[below] {$y$} (1,0);
 \draw[<->] (1.5,0) -- node[below] {$x$} (2.5,0);
 \end{tikzpicture} &
 \begin{tikzpicture}
\draw[<->] (0,0) -- node[below] {$y$} (1,0);
 \draw[<->] (1,0) -- node[below] {$x$} (2,0);
 \end{tikzpicture} &
 \begin{tikzpicture}
 \draw[<->] (0,0) -- node[below] {$y$} (1.5,0);
 \draw[<->] (1,0) -- node[below] {$x$} (2.5,0);
 \end{tikzpicture}\tabularnewline
starts $s$ & during $d$ & finishes $f$\tabularnewline
\begin{tikzpicture}
\draw[<->] (0,0) -- node[below] {$x$} (0.75,0);
 \draw[<->] (0,0) -- node[below] {$y$} (2,0);
 \end{tikzpicture} &
 \begin{tikzpicture}
\draw[<->] (0.25,0) -- node[below] {$x$} (1,0);
 \draw[<->] (0,0) -- node[below] {$y$} (2,0);
 \end{tikzpicture} &
 \begin{tikzpicture}
 \draw[<->] (1.25,0) -- node[below] {$x$} (2,0);
 \draw[<->] (0,0) -- node[below] {$y$} (2,0);
 \end{tikzpicture}\tabularnewline
started by $si$ & contains $di$ & finished by $fi$\tabularnewline
\begin{tikzpicture}
\draw[<->] (0,0) -- node[below] {$y$} (0.75,0);
 \draw[<->] (0,0) -- node[below] {$x$} (2,0);
 \end{tikzpicture} &
 \begin{tikzpicture}
\draw[<->] (0.25,0) -- node[below] {$y$} (1,0);
 \draw[<->] (0,0) -- node[below] {$x$} (2,0);
 \end{tikzpicture} &
 \begin{tikzpicture}
 \draw[<->] (1.25,0) -- node[below] {$y$} (2,0);
 \draw[<->] (0,0) -- node[below] {$x$} (2,0);
 \end{tikzpicture}\tabularnewline
\multicolumn{3}{c}{equals $eq$}\tabularnewline
\multicolumn{3}{c}{ \begin{tikzpicture}
 \draw[<->] (0,0) -- node[below] {$x = y$} (1.5,0);
  \draw[<->] (0,0) -- (1.5,0);
 \end{tikzpicture}}
\end{tabular}
\caption{A representation of the $13$ base relations $b$ of  Interval Algebra ($\IA$)~\cite{DBLP:journals/cacm/Allen83}\label{figA:A}, each one relating two potential intervals $x$ and $y$ as in $x~b~y$; here, ${bi}$ denotes the converse of $b$, i.e., $b^{-1}$.}
\end{figure}
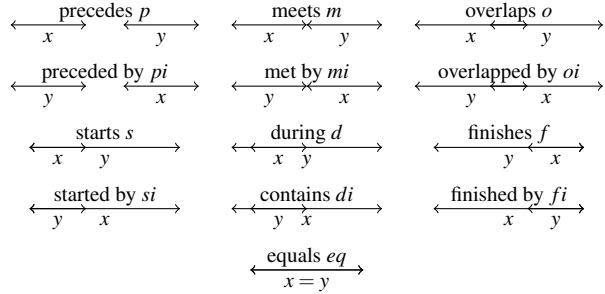

As an illustration, consider the well-known qualitative temporal constraint language of Interval Algebra ($\IA$), introduced by Allen~\cite{DBLP:journals/cacm/Allen83}. $\IA$ considers time intervals (as temporal entities) and the set of base relations $\B=\{eq$, $p$, $pi$, $m$, $mi$, $o$, $oi$, $s$, $si$, $d$, $di$, $f$, $fi\}$ to encode knowledge about the temporal relations between intervals on the timeline, as described in Figure~\ref{figA:A}.
\begin{figure}[h!]
\centering
\scriptsize
\strut
{
\def\texteX{\text{\tiny{$X$}}}
\def\texteY{\text{\tiny{$Y$}}}
\pgfmathsetmacro{\grandR}{.5}
\pgfmathsetmacro{\moyenR}{.3}
\pgfmathsetmacro{\petitR}{.2}
\begin{tabular}{cccccc}
  \begin{tikzpicture}
    \coordinate (X) at (0, 0) ;
    \coordinate (Y) at (2 * \moyenR + .2, 0, 0) ;
    \draw (X) circle (\moyenR) ;
    \draw (Y) circle (\moyenR) ;
    \node[right=-3mm of X]{\texteX} ;
    \node[left=-3mm of Y]{\texteY} ;
  \end{tikzpicture}
&
  \begin{tikzpicture}
    \coordinate (X) at (0, 0) ;
    \coordinate (Y) at (2 * \moyenR, 0) ;
    \draw (X) circle (\moyenR) ;
    \draw (Y) circle (\moyenR) ;
    \node[right=-3mm of X]{\texteX} ;
    \node[left=-3mm of Y]{\texteY} ;
  \end{tikzpicture}
&
  \begin{tikzpicture}
    \coordinate (X) at (0, 0) ;
    \coordinate (Y) at (2 * \moyenR - 0.2, 0) ;
    \draw (X) circle (\moyenR) ;
    \draw (Y) circle (\moyenR) ;
    \node[right=-3.5mm of X]{\texteX} ;
    \node[left=-3.5mm of Y]{\texteY} ;
  \end{tikzpicture}
&
  \begin{tikzpicture}
    \coordinate (X) at (0, 0) ;
    \draw (X) circle (\moyenR) ;
    \node[right=-3.5mm of X]{\texteX} ;
    \node[left=-3.5mm of X]{\texteY} ;
  \end{tikzpicture}
&
  \begin{tikzpicture}
    \coordinate (Y) at (0, 0) ;
    \coordinate (X) at (-\petitR, -\petitR) ;
    \draw (Y) circle (\grandR) ;
    \draw (X) circle (\petitR) ;
    \node[right=0mm of Y]{\texteY} ;
    \node[left=-2.5mm of X]{\texteX} ;
  \end{tikzpicture}
&
  \begin{tikzpicture}
    \coordinate (Y) at (0, 0) ;
    \coordinate (X) at (-\petitR+.1, -\petitR+.1) ;
    \draw (Y) circle (\grandR) ;
    \draw (X) circle (\petitR) ;
    \node[right=0mm of Y]{\texteY} ;
    \node[left=-2.5mm of X]{\texteX} ;
  \end{tikzpicture}
\tabularnewline
$X~DC~X$ & $X~EC~Y$ & $X~PO~Y$ & $X~EQ~X$ & \makecell{$X~TPP~Y$\\$Y~TPPi~X$} & \makecell{$X~NTPP~Y$\\$Y~NTPPi~X$}
\end{tabular}
}
\caption{The base relations \calc{RCC8}.}
\label{fig:RCC8}
\end{figure}
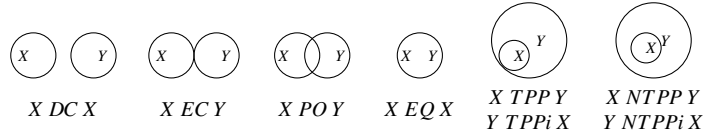

As another example, the Region Connection Calculus (\calc{RCC8})~\cite{randell1992spatial} considers spatial regions and the set of base relations $\B=\{DC$, $EC$, $EQ$, $PO$, $TPP$, $TPPi$, $NTPP$, $NTPPi \}$ to reason about topological relations between regions, as can be seen in Figure~\ref{fig:RCC8}.

\begin{figure}[ht!]
  \centering
  \includegraphics[width=0.3\linewidth]{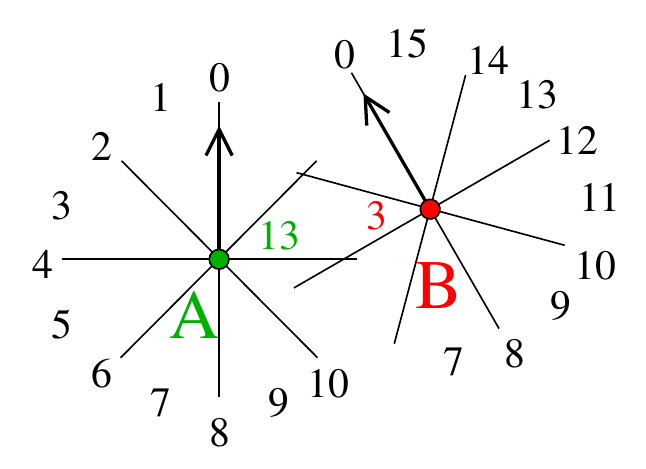}
  \caption{An $\calc{OPRA}_4$ relation $A {}_{4}\angle_{13}^{3} B$. Numbers $3$ and $13$ in $A {}_{4}\angle_{13}^{3} B$ correspond to the sector numbers of the entities, and number $4$ stands for the number of lines used to obtain the sectors.}
  \label{fig:opra4}
\end{figure}

Relative directions~\cite{lee_complexity_2014,mossakowski_qualitative_2012,wolter_qualitative_2010} are commonly used to refer to a target object from the perspective of a reference object (e.g., ``A cat in \emph{front} of a car.''). For representing and reasoning with relative directions, one can for example use the constraint language $\calc{OPRA}_m$~\cite{mossakowski_qualitative_2012}, which relates two oriented points with respect to their relative positions and orientations (cf.~Figure~\ref{fig:opra4}).

Finally, the challenge of representing and reasoning about qualitative spatio-temporal information can be facilitated
by a \emph{qualitative spatio-temporal network}, for which we recall the following definition:

\begin{definition}\label{def:qcn}A \emph{qualitative spatio-temporal network} is a tuple $(V,C)$ where: 
\begin{itemize}
  \item $V$ $=$ $\{v_1,$ $\ldots,$ $v_n\}$ is a finite set of variables over some infinite domain $\D$ (e.g., time points or 2D regions
);
\item and $C$ is a mapping $C:V\times{V}\rightarrow{2^\B}$ associating a relation (set of base relations) with each pair of variables.
\end{itemize}
\end{definition}

A simplified example of a qualitative spatio-temporal network can be seen on the right-hand side of Figure~\ref{fig:simpleqcn}.


\section{Neuro-Symbolic Spatio-Temporal Reasoning and Learning}

In this section, we introduce some works on neuro-symbolic spatio-temporal reasoning and learning (Sections~\ref{sec:relwork} and \ref{sec:symb-meta-learn}) and propose a generic neuro-symbolic framework for integrating qualitative spatio-temporal reasoning and neural methods from a probabilistic perspective (Section~\ref{sec:prob}). As datasets are crucial for training and evaluating such neuro-symbolic frameworks, we also review in this section a selection of recent datasets for spatial relation learning (Section~\ref{sec:datasets}). We start this section by first motivating why spatio-temporal logic is important for an intelligent system.
\label{sec:neuro-symb-spat}

\subsection{Logic vs. Spatio-Temporal Logic}
\label{sec:stlogic}

Within a Neuro-Symbolic AI framework, any type of logic can be used, such as inductive logic, deductive logic, abductive logic, some flavor of modal logic, and so on. Typically, machine learning itself can be viewed as a form of inductive reasoning, where a general rule is learned from a set of observations with the aim of developing a theory. Complementing the advantages of machine learning by having logic at the other end either to verify the theory via deductive reasoning or to translate the derived theory into a new sound understanding via abductive reasoning can be beneficial in many different contexts.

However, what is more important than the type of logic to be used in a neuro-symbolic framework is the application domain of the logic, i.e., the type of information the logic is dealing with. Let us consider the following statement, inspired by a similar set of examples as found in~\cite{DBLP:journals/pieee/ScholkopfLBKKGB21}:
\begin{center}
\emph{``If I accelerate faster than the vehicle right in front of me, then I will overtake it.''}
\end{center}
The above statement, even though it forms a logic clause, is not logical, primarily because it does not conform to the laws of physics and human cognition. Let us contrast it with the following statement:
\begin{center}
\emph{``If I accelerate faster than the vehicle right in front of me, then I will bump into it.''}
\end{center}
The above statement forms a rule that is both logical and adheres to the laws of physics and human cognition. 
Thus, it is a rule that encodes \emph{hard} and verifiable knowledge that an intelligent system should
inherently possess, or learn through its interaction with the physical environment. By extension, we argue that spatio-temporal logic, which is naturally grounded in physics and human cognition,
forms a minimal requirement for an intelligent system to be able to generalize in a causally sound manner.

Indeed, since spatio-temporal formalisms encode valid 
assumptions that align with our understanding of physics and human cognition, a learner that does not use them should fare worse than
one that does. Interestingly, it is argued that even for humans certain spatial relations, such as the connection relation, are primitive/primordial, and are crucial for the human brain to be able
to generalize and develop a better understanding of the spatio-temporal dynamics in its environment~\cite{DBLP:series/sci/2011-388}.



\subsection{Example 1: A Semantic Referee for Geospatial Semantic Segmentation of Satellite Imagery Data}
\label{sec:relwork}

\begin{figure*}[t]
\centering
\includegraphics[width=\textwidth]{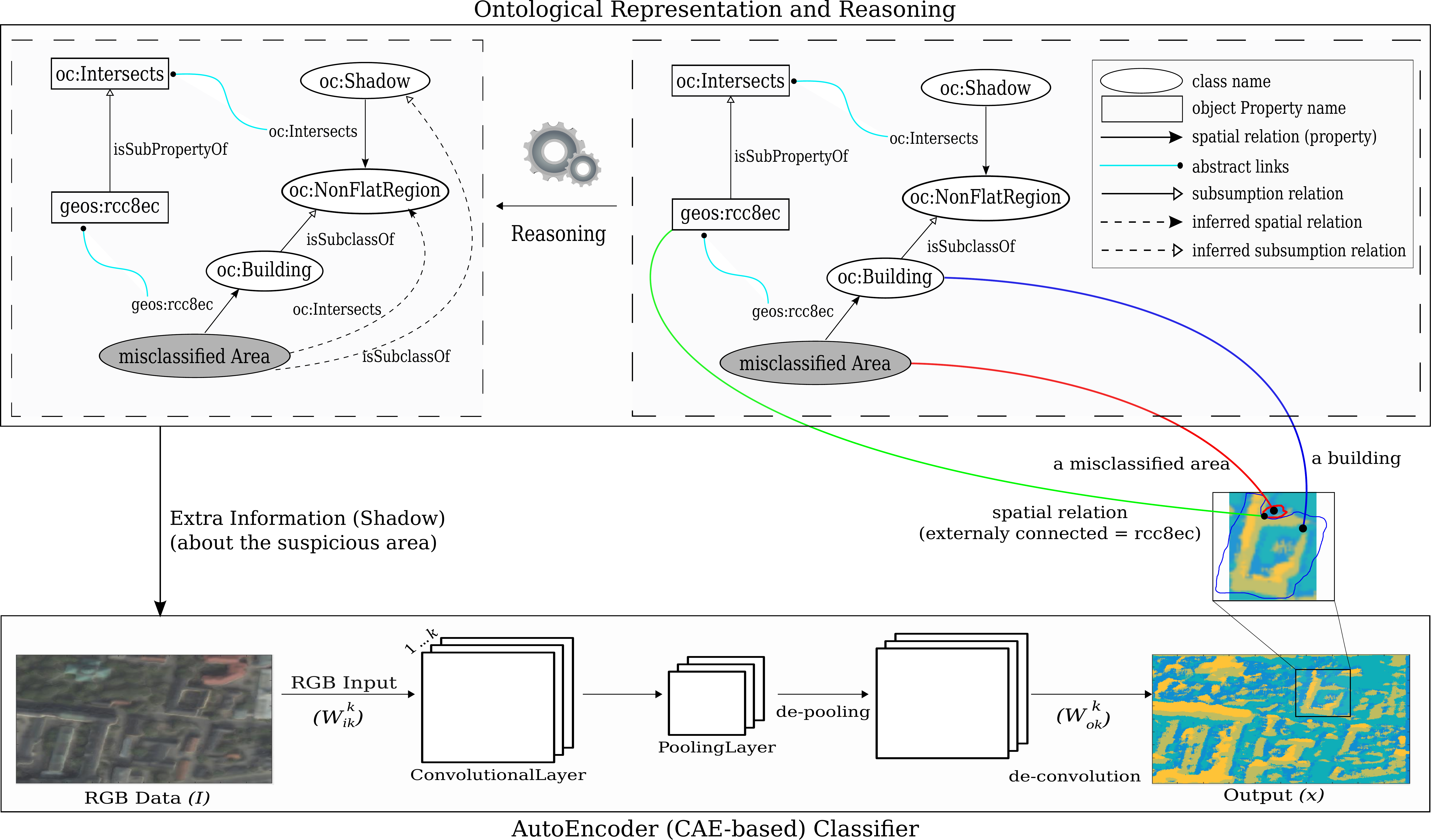}
\caption{Overview of applying a semantic referee (top layer) in the form of reasoning upon ontological knowledge to improve the semantic segmentation task of the convolutional encoder-decoder classifier (bottom layer). The semantic referee reasons about the mistakes made by the classifier based on ontological concepts and provides additional information back to the classifier that prevents the classifier from making the same misclassifications. The ontological knowledge and reasoning methods play the role of the semantic referee that makes sense of the errors from the classifier.}
\label{fig:flow-1}
\end{figure*}


In~\cite{DBLP:journals/semweb/AlirezaieLSL19}, a neuro-symbolic approach is proposed to ameliorate the geospatial semantic segmentation of satellite imagery data. 
An overview of this approach can be seen in Figure~\ref{fig:flow-1}, which demonstrates the interaction between a classifier and a logic-based component that is termed as a \emph{semantic referee}. 
Specifically, the classifier here is a deep convolutional network with an encoder-decoder structure that provides the semantic segmentation of the input image.
With respect to the logic-based component, \calc{RCC8} spatial relations are used (as well as some extensions thereof), which we introduced in Section~\ref{sec:qstr}. In addition to the mentioned works, \calc{RCC8} has also been adopted by the GeoSPARQL\footnote{\url{http://www.opengeospatial.org/standards/geosparql}} standard and has found its way into various Semantic Web tools and cross-disciplinary applications over the past few years~\cite{DBLP:conf/rweb/KoubarakisKKNS12}. As an example of such an application, certain spatial facts encoded in \calc{RCC8}, e.g., that a nucleus should be properly contained in a cell, have been used to correct segmentation errors for images of hematoxylin and eosin (H\&E)-stained human carcinoma cell line cultures~\cite{DBLP:journals/jimaging/RandellGFML17}. In ~\cite{DBLP:journals/semweb/AlirezaieLSL19}, inspired by the integration of qualitative spatial reasoning methods into imaging procedures as described by Randell et al., 2017~\cite{DBLP:journals/jimaging/RandellGFML17}, we aim to employ spatially-enhanced ontological reasoning techniques in order to assist deep learning methods for image classification via interaction and guidance. Thus,
in order to deal with the misclassifications from the classifier, \calc{RCC8} reasoning, acting as a semantic referee, reasons about the errors that include the conceptualization of the misclassified regions based on their physical (e.g., geometrical) properties and aligns them with the available ontology to infer the best possible match for the error.
The inferred concept related to the misclassified region is then given to the classifier as a referee, providing information to be used within the learning process to prevent the classifier from making the same misclassifications. The additional information provided by the reasoner is represented as image channels with the same size as the RGB input and is then concatenated together with the original RGB channels as additional color channels. In this work, three additional channels from the reasoner are used that represent shadow estimation, elevation estimation, and other inconsistencies respectively. This results in the classifier using data with six channels instead of the original three RGB channels.
This process is then repeated until the classification accuracy on the validation data converges. During testing, the same procedure is performed using the same number of iterations that was used during training.

A major shortcoming of the above approach, which we address in this chapter and in Section~\ref{sec:prob} in particular, is that it assumes the given knowledge base to be accurate and
consistent. Thus, the knowledge base acts as an expert (or, referee) that is not subject to any form of revision. However, in practical real-life settings, the available knowledge base may be inaccurate/uncertain or may contain contradicting clauses (i.e., it can be inconsistent). Consequently, the knowledge base itself should
be subject to refinement. This would involve optimization on symbolic relations rather than numerical optimization, which is a technical challenge that has not been explored much in the context of QSTR. If we revisit the previous example, and Figure~\ref{fig:flow-1} in particular, one can imagine many cases where the dark region externally connected to a tall building is indeed water and not shadow, as the semantic referee would invariably claim. In addition to erroneously identifing/labeling regions, mistakes can also occur in the identification of the semantic/spatial relations between regions because of, for example, poor image resolution or insufficient pixel analysis (cf.~\cite{DBLP:conf/ifip12/SioutisCSMR15}). 

\subsection{Example 2: A Symbolic Meta-Learning Interface for Learning Relative Directions}
\label{sec:symb-meta-learn}

A complex reasoning task often consists of subtasks that can be solved independently. For example, given a scene and a question about the objects in the scene, answering questions about the relative directions between objects comprises intuitively the following subtasks\ms{:}
\begin{inparaenum}[(i)]
  \item object detection
  \item orientation estimation, and
  \item relation identification.
\end{inparaenum}
A common approach to solving a complex task is building a pipeline of modules, where each module is responsible for solving a subtask. For example, to ground relative directions, one would need three modules corresponding to the three subtasks mentioned above. Such modules are often handcrafted. However, there are powerful end-to-end differentiable neural models that can be applied to many different downstream tasks~\cite{lu_12--1_2020,wang_ofa_2022}. A question is how to train such powerful neural models so that they can solve complex tasks such as the one mentioned above.

We argue that for such universal models, we can shift our focus from devising a model for the complex task to \emph{curating a dataset} that can guide the model to solve the complex task. The dataset is generated and curated manually by a domain expert. This kind of neuro-symbolic learning and reasoning is related to the hybrid reasoning pattern \emph{meta-reasoning for control} as proposed in \cite[Figure~11]{van_bekkum_modular_2021}. The difference is, however, that the focus is more on solving a complex task with a symbolic curation of a multi-task dataset and not on symbolic intervention for improving the models (e.g., AutoML) or the training setup (e.g., curriculum learning).

\begin{figure*}[t]
  \hfill%
  \subfloat[Overview of the eight different relative directions from the viewpoint of the reference object \textit{elephant}.]{%
    \centering%
    \includegraphics[width=0.4\linewidth]{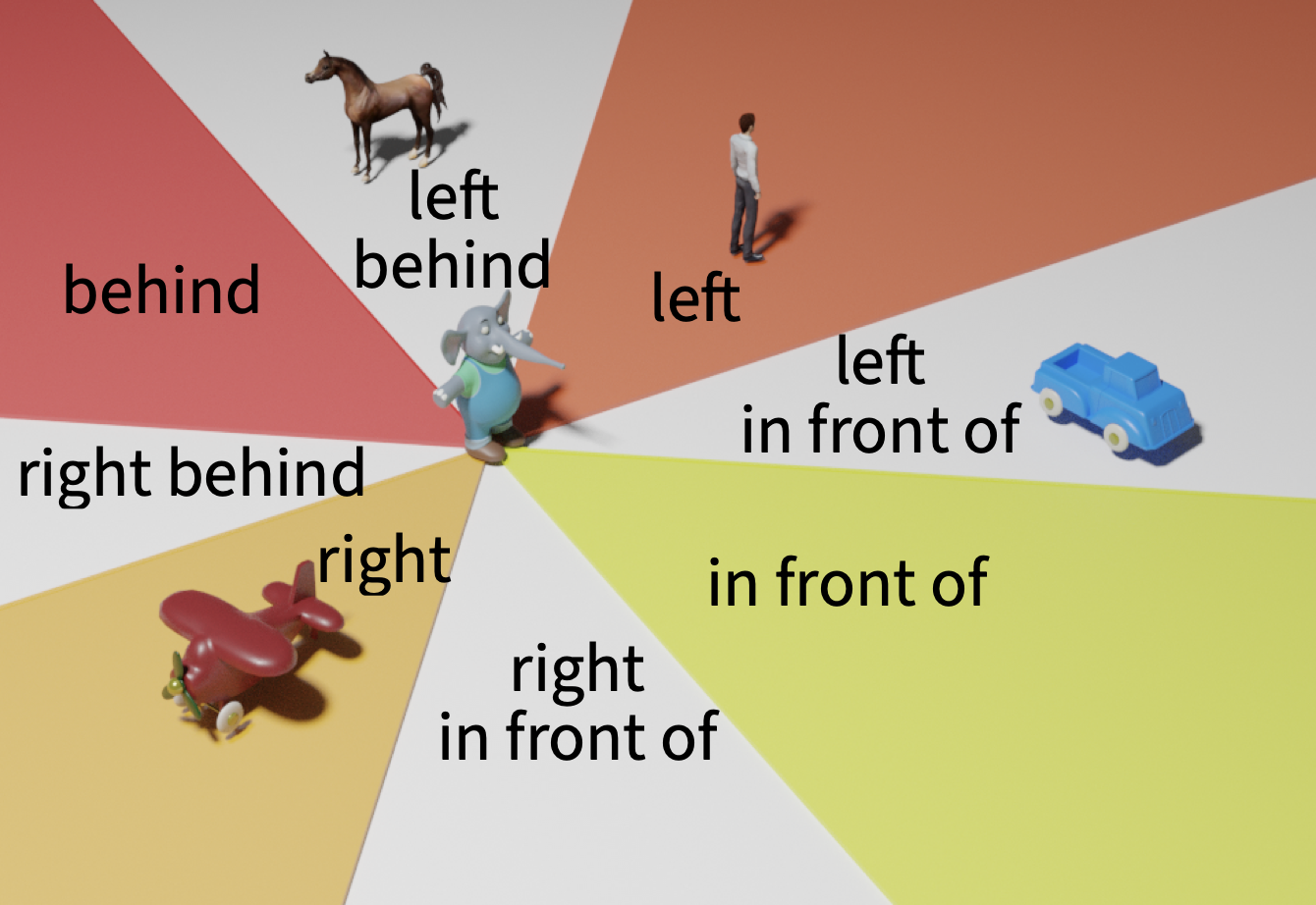}%
  }%
  \hfill%
  \subfloat[All 28 3D-objects used for the synthetic dataset. Each object features a clear front side.]{%
    \centering%
    \includegraphics[width=0.425\linewidth]{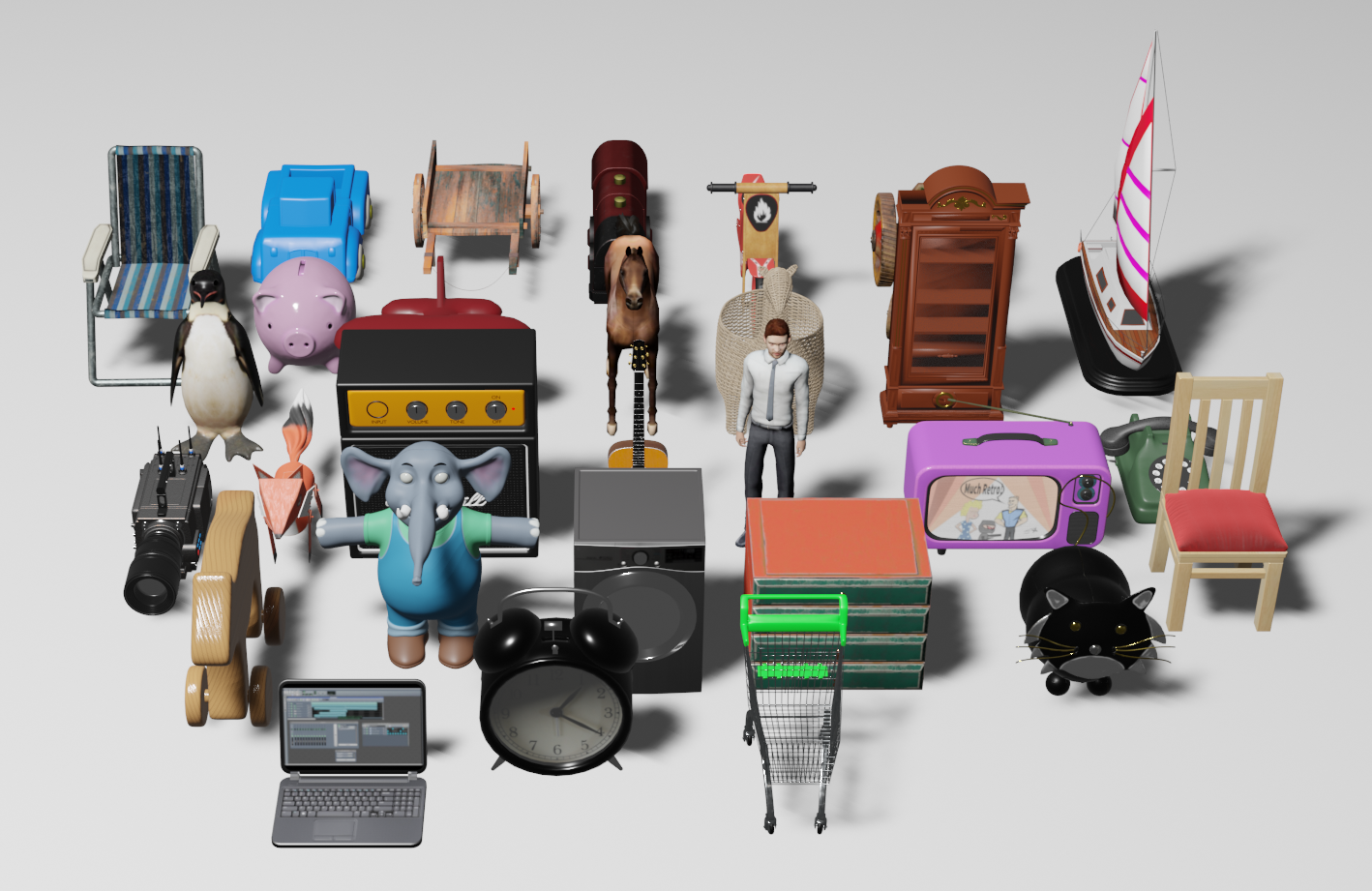}%
  }%
  \hfill \phantom{.}%
  \caption{The GRiD-3D dataset.}
  \label{fig:taskSummaryOverview}
\end{figure*}

In \cite{lee_what_2022}, we introduce a visual question answering (VQA) dataset, called GRiD-3D, for learning relative directions, where the task is to answer a question about an image (see~Figure~\ref{fig:taskSummaryOverview}). A special feature of the dataset is that the symbolic knowledge about the necessary three steps above (i.e., object detection, orientation estimation, and relation identification) to learn relative directions is injected into the dataset as suggested previously. To be precise, the dataset consists of the six different tasks in Table~\ref{tab:grid3d}. Note that the first task \task{Existence Prediction} is a prerequisite for solving the second task \task{Orientation Prediction}. Similarly, the second task \task{Orientation Prediction} is a prerequisite for the remaining four tasks \task{Link Prediction}, \task{Relation Prediction}, \task{Counting}, and \task{Triple Classification}, which require knowledge about relative directions.

\begin{table}[t]
  \caption{The six tasks in the GRiD-3D dataset.\label{tab:grid3d}}
  \centering
  \begin{tabular}{lll}
    Task name & Example & Answer set\\
    \hline
    \task{Existence Prediction} & ``Is there $x$ in the image?'' & \{yes, no\}\\
    \task{Orientation Prediction} & ``What is the orientation of $x$?'' & \{north, east, south, west\} \\
    \task{Link Prediction} & ``What is to the \emph{left} of $x$?'' & 28 object names\\
    \task{Relation Prediction} & ``What is the relation between $x$ and $y$?'' & 8 relations \\
    \task{Counting} & ``How many objects are to the \emph{right} of $x$?'' & 2--5\\
    \task{Triple Classification} & ``Is $x$ in \emph{front} of $y$?'' & \{yes, no\}
  \end{tabular}
\end{table}

To show the effect of the symbolic intervention in the dataset, we train the two well-known VQA models FiLM~\cite{perez_film_2018} and MAC~\cite{hudson_compositional_2018} on the six tasks of the GRiD-3D dataset in a multi-task learning setup. In particular, we are interested in answering the following questions:
\begin{enumerate}
  \item Is multi-task learning necessary for grounding relative directions?
  \item Is multi-task learning with GRiD-3D sufficient to ground relative directions?
  \item Is there an order in learning the six tasks?
\end{enumerate}

\begin{figure}[t]
  \centering%
  \hfill%
  \subfloat[%
  Trained on \task{Triple Classification} only.\label{fig:single-task}%
  ]{%
    \centering%
    \includegraphics[width=0.4\linewidth]{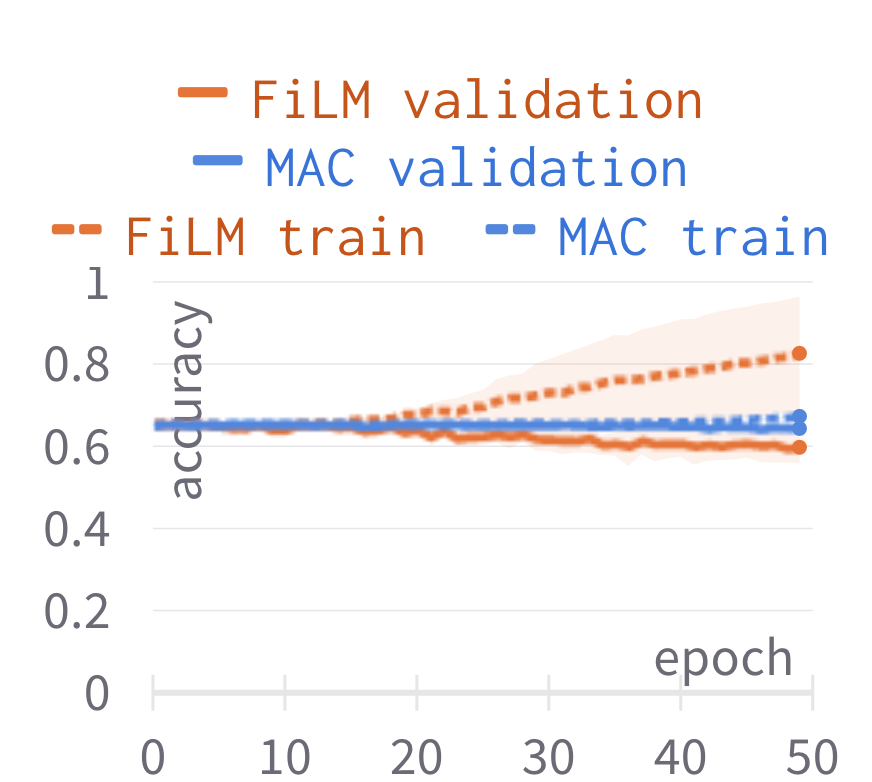}%
  }%
  \hfill%
  \subfloat[%
  Trained on all tasks in GRiD-3D.\label{fig:multi-task}%
  ]{%
    \centering \includegraphics[width=0.4\linewidth]{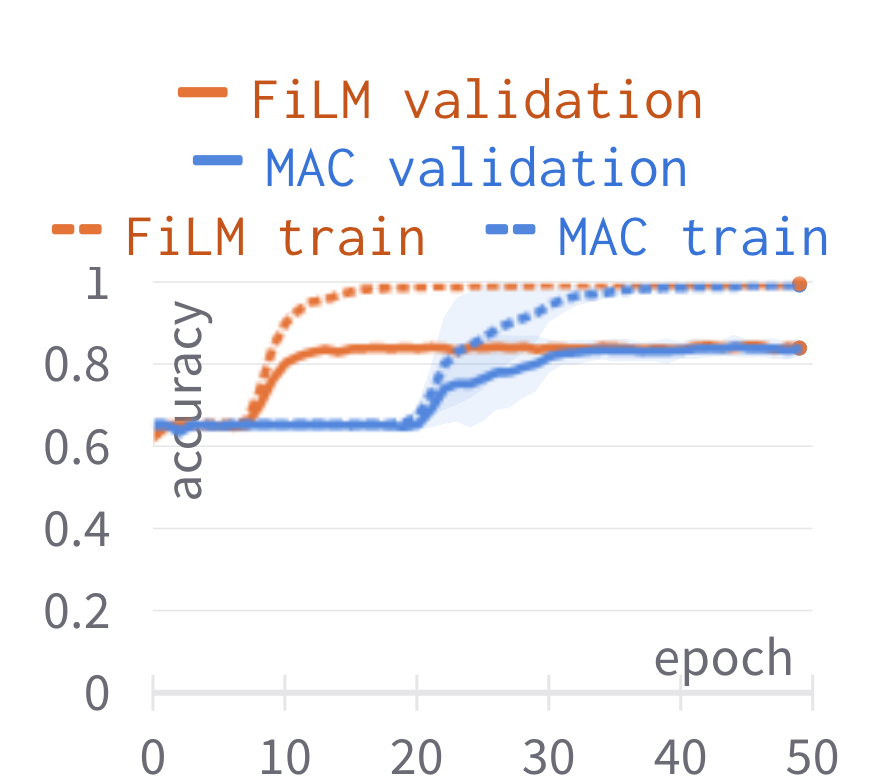}%
  }%
  \hfill%
  \phantom{.}
  \caption{Mean validation and training accuracies of FiLM and MAC on \task{Triple Classification}. The mean accuracies are obtained by averaging three runs.}
\end{figure}

\paragraph{Multi-Task Learning is Necessary.}
To answer the first question, we first trained the two models on the \task{Triple Classification} task (cf.~Figure~\ref{fig:single-task}), which involves relative directions in the questions and thus requires the models to ground relative directions. When the models were trained only on the \task{Triple Classification} task, they were not able to learn to solve the task (cf.~the training curve of MAC in Figure~\ref{fig:single-task}), which suggests that training on a single task is not sufficient, but multiple tasks are necessary to solve \task{Triple Classification}.

\paragraph{Multi-task Learning with GRiD-3D is Sufficient.}
Although we showed the necessity of multi-task learning, it is not clear whether the six tasks in the GRiD-3D dataset are sufficient to allow the models to ground relative directions. To investigate this, we trained FiLM and MAC on all six GRiD-3D tasks and evaluated their validation performance. Figure~\ref{fig:multi-task} shows that both the training and the validation accuracies on the \task{Triple Classification} show significant improvements when trained on all six tasks. First, the models were able to learn the training data and achieve about 100\% accuracy, which was not possible when trained on \task{Triple Classification} only. Second, the models achieve higher performance on the validation set of \task{Triple Classification}. Given that there are difficulties in detecting objects (e.g., due to occlusions), estimating orientations (e.g., due to almost symmetric objects)\ms{,} and identifying relations (e.g., target objects being close to the decision boundary between two relations), further improving the performance would be challenging without injecting additional prior knowledge to the models. We can therefore conclude that the tasks in GRiD-3D are sufficient to allow FiLM and MAC to solve the \task{Triple Classification} task.

\paragraph{The Order of Learning the Tasks.}

\begin{figure*}[t]
  \subfloat[FiLM]{%
    \begin{minipage}[t]{0.4\linewidth}
      \centering%
      \includegraphics[width=1.0\linewidth]{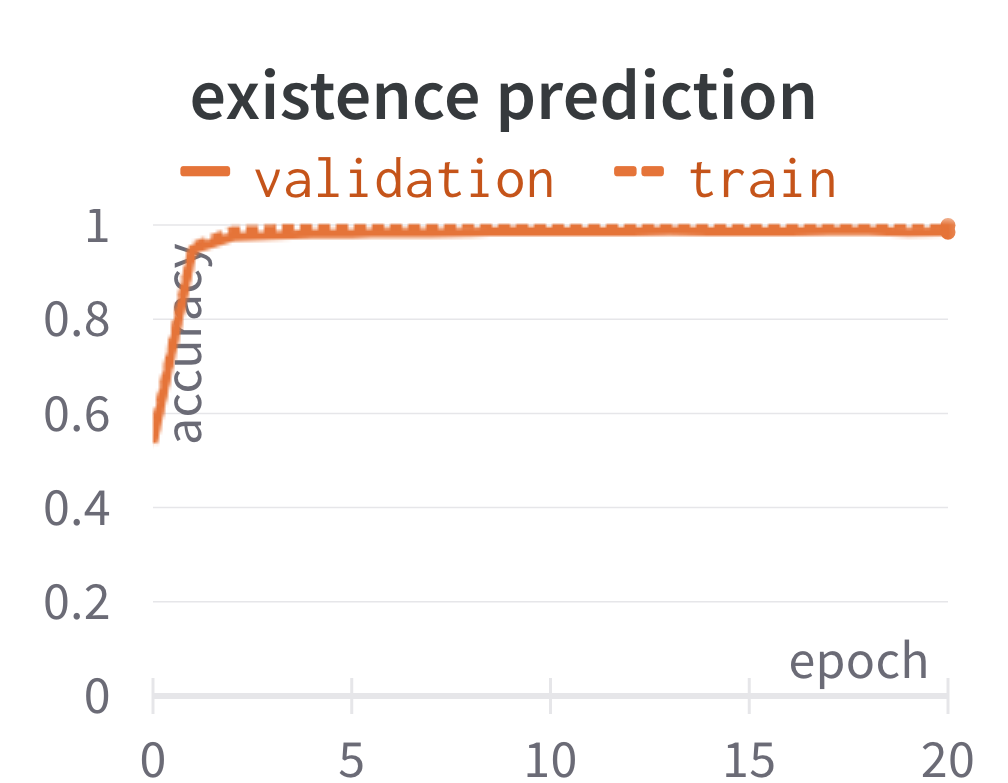}\\%
      \includegraphics[width=1.0\linewidth]{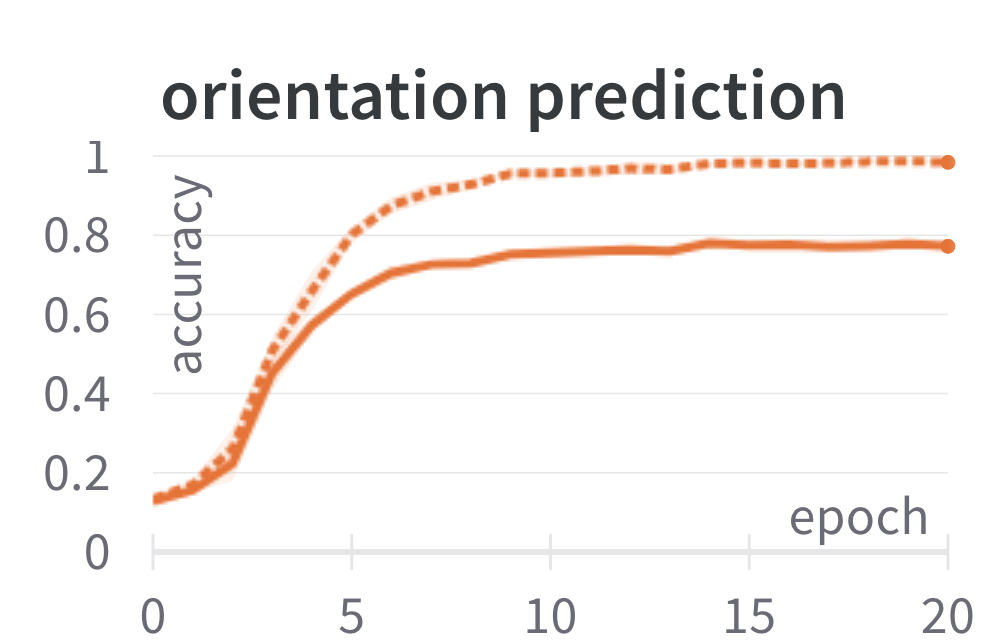}\\%
      \includegraphics[width=1.0\linewidth]{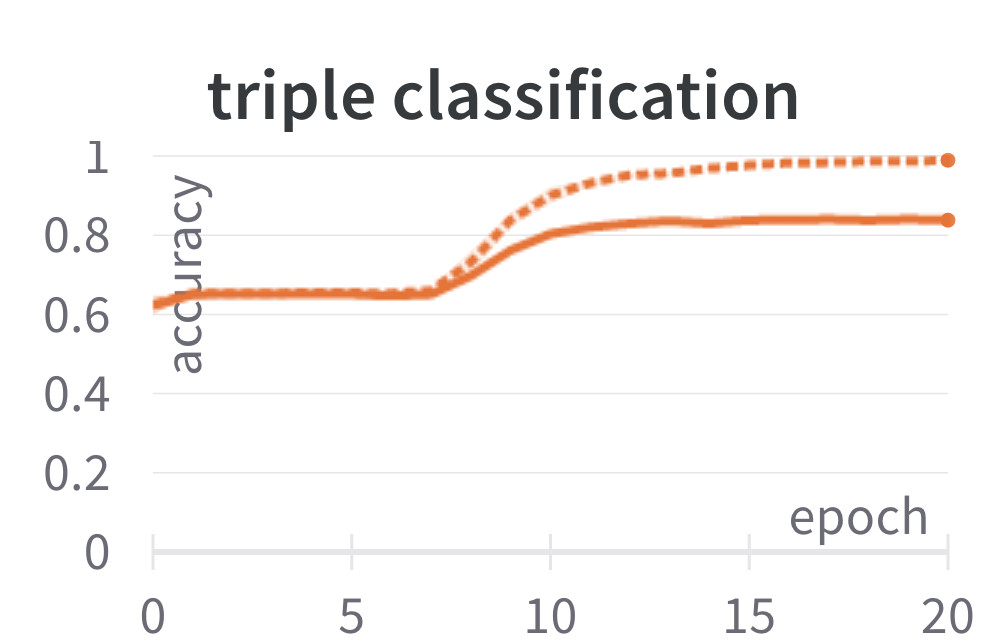}
    \end{minipage}
  }%
  \subfloat[MAC]{%
    \begin{minipage}[t]{0.4\linewidth}
      \centering%
      \includegraphics[width=1.0\linewidth]{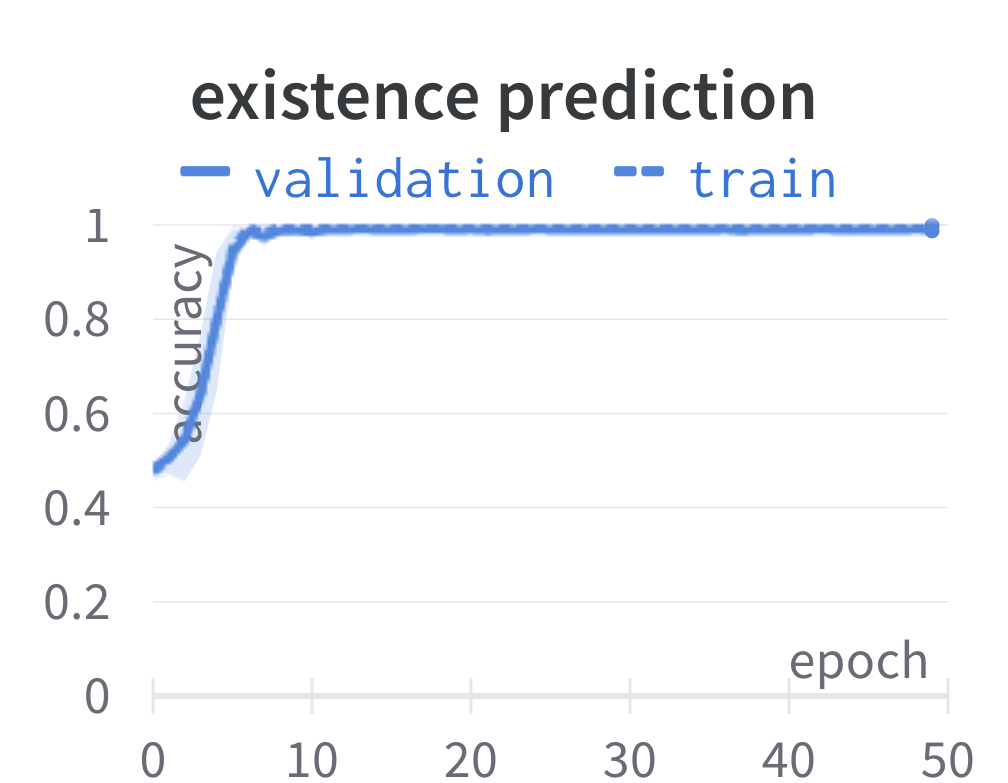}\\%
      \includegraphics[width=1.0\linewidth]{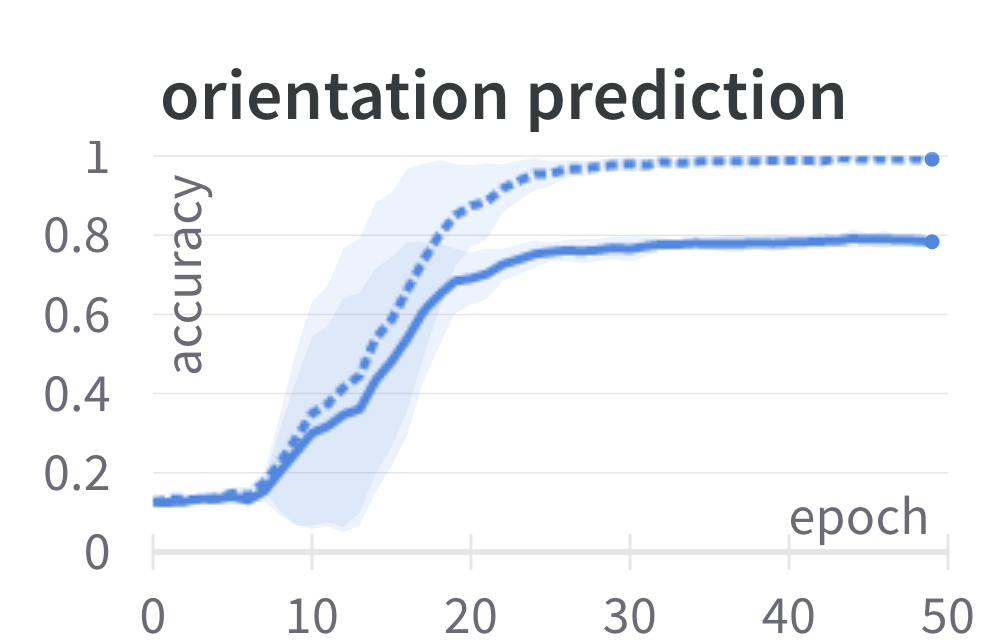}\\%
      \includegraphics[width=1.0\linewidth]{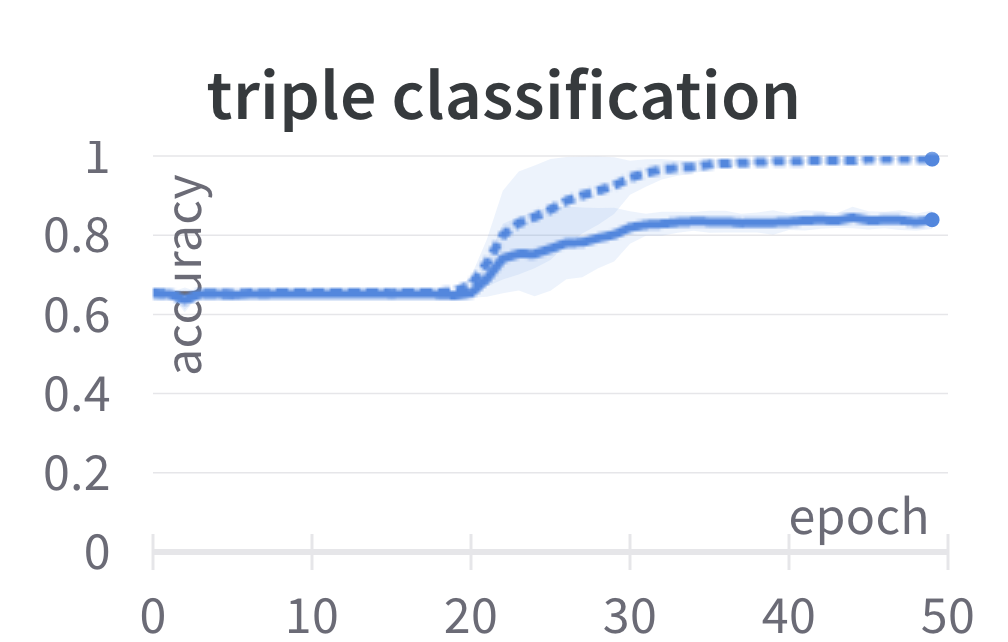}
    \end{minipage}
  }%
  \caption{Multi-task learning results of FiLM and MAC on \task{Existence Prediction}, \task{Orientation Prediction} and \task{Triple Classification}. One can observe that there is an order in learning the tasks reflecting our intuition discussed at the beginning of Section~\ref{sec:symb-meta-learn}
  \label{fig:learning-dynamics}}
\end{figure*}

At the beginning of Section~\ref{sec:symb-meta-learn}, we discussed an intuitive order of subtasks for grounding relative directions, i.e.,
\begin{inparaenum}[(i)]
  \item object detection\ms{,}
  \item orientation estimation, and
  \item relation identification.
\end{inparaenum}
A natural question to ask is whether this order is reflected in the learning dynamics of the two models FiLM and MAC when evaluated on the GRiD-3D dataset. The results shown in Figure~\ref{fig:learning-dynamics} indicate that the intuitive order in learning the subtasks is reflected in the learning dynamics: First, \task{Existence Prediction} is learned, followed by \task{Orientation Prediction}, and finally, \task{Triple Classification}. This emergence of an intuitive but surprising implicit order could give an answer to the question of \emph{why} and \emph{how} adding more tasks (while keeping the same images in the training set) leads to an improvement both in the training and the validation performances. A complex task in GRiD-3D, such as \task{Triple Classification}, is dependent on the knowledge of simpler tasks, such as \task{Existence Prediction} and \task{Orientation Prediction}. In other words, a model has to learn the latter simpler tasks first before solving the more complex task \task{Triple Classification}.

In summary, analyzing our experimental results reveals that curating a dataset based on a prior symbolic knowledge of the target task can help neural models solve complex tasks, which in particular allows the two VQA models to ground relative directions in the GRiD-3D dataset.


\subsection{Outlook: Probabilistic Infusion}
\label{sec:prob}

In~\cite{DBLP:journals/semweb/AlirezaieLSL19}, and as also summarized in Section~\ref{sec:relwork}, QSTR acts as a model-based referee upon the output of a classifier to improve the performance of semantic segmentation for satellite imagery data by rejecting impossible interpretations. Drawing inspiration from that work, we argue for working towards a \emph{generic} neuro-symbolic framework that will integrate QSTR and neural methods from a probabilistic perspective, which can be particularly important in the context of uncertain environments (see Section~\ref{sec:relwork}). Such an integration is currently identified as an open challenge in the AI community~\cite[Section~$9$]{DBLP:conf/ijcai/RaedtDMM20}, yet it is not clear what current and novel QSTR techniques are needed to empower handling spatio-temporal information.

Providing a neuro-symbolic framework with the ability to handle uncertainty is also motivated in prominent works in the literature, such as~\cite{DBLP:journals/ai/SuchanBV21}.
In that work, a general neuro-symbolic method for online \emph{visual sensemaking}\footnote{This is an umbrella term for any modular reasoning framework where visual perception is needed as one of the several components.} using Answer Set Programming (ASP) is formalized and implemented, as well as showcased in the context of autonomous driving.
The method builds upon a tight but modular integration of (declarative) commonsense spatio-temporal abstraction and reasoning with robust low-level visual computing foundations (primarily) driven by state-of-the-art visual computing techniques (e.g., for visual feature detection, and tracking), and is generally used within hybrid architectures for real-time perception and control; for the spatio-temporal aspect in particular, an extension of Interval Algebra is used among other representations~\cite{Guesgen89RectangleAlgebra}. However, 
this work handles the uncertainty involved in low-level object tracking in a very limited manner, i.e., a full-scale systematic formalization of a probabilistic model has not been attempted therein, as argued by the authors. 


Our first approach to such a neuro-symbolic framework appears in~\cite{swi21}. Specifically, in~\cite{swi21} we make the case for neurally enhanced QSTR by proposing to use probabilistic values to construct a \emph{bidirectional feedback loop} between the ML model and the symbolic framework in an abductive reasoning-based manner that is referred to in the recent literature as \emph{abductive learning}~\cite[Figure~$1$]{DBLP:journals/chinaf/Zhou19}; see also~\cite{DBLP:conf/nips/DaiX0Z19}. In particular, we propose to annotate the variables in a qualitative spatio-temporal configuration with the probabilistic values produced by the ML model, and likewise assign probabilistic values to the constraints of the spatio-temporal configuration. A neurally enhanced symbolic relation could then be as follows ($\%$s denote likelihood):
$$
X^{(95\%~\text{yolk})}~\text{is}~\text{\emph{contained in}}^{(45\%~\text{true})}~\text{or}~\text{\emph{overlaps}}^{(55\%~\text{true})}~Y^{(90\%~\text{egg})}
$$

Here, we assume that the neural network has identified variables $X$ and $Y$ to be yolk and egg with $95\%$ and $90\%$ confidence, respectively (e.g., by means of image recognition). Likewise, it was calculated that relations \emph{contained} and \emph{overlaps} hold with $45\%$ and $55\%$ confidence respectively (e.g., via image segmentation or human annotation) due to some ambiguity involving the boundaries of the regions (e.g., because of insufficient image resolution).
Next, we can consult some background knowledge (e.g., that a yolk is contained in an egg) and let the classifier rectify the spatial knowledge base, either by increasing the confidence of relation \emph{contained} or getting rid of relation \emph{overlaps} altogether (towards maximizing consistency). Likewise, there could be cases, as in~\cite{DBLP:journals/semweb/AlirezaieLSL19}, where the QSTR knowledge base ameliorates the output of a classifier, but, in contrast to~\cite{DBLP:journals/semweb/AlirezaieLSL19}, does that probabilistically (e.g., by modifying weight parameters across neural network layers or simply enriching the input of the classifier). This is the type of collaborative probabilistic reasoning between ML and Symbolic AI that we want to achieve, where logic is used to compose (partial) knowledge learned by ML methods, and this learned knowledge may in turn influence that logical composition.


To this end, it is important to establish probabilistic robustness measures relating to QSTR frameworks, and also construct adaptive and dynamic algorithms for verifying neural network-based components at runtime. In~\cite{DBLP:conf/iske/SioutisM21}, we build upon the work in~\cite{ijcai2020} and take the next step by proposing such a measure as follows:

Given a qualitative spatio-temporal network $\mathcal{N}=(V,C)$, an atomic refinement ${\mathcal N}^\prime$ of ${\mathcal N}$, i.e., a subnetwork where only a single base relation of ${\mathcal N}$ is used for each of its constraints, and the probability $p_{ij}({\mathcal N}^\prime[v_i,v_j])$ of a relation ${\mathcal N}^\prime[v_i,v_j]$ appearing in any scenario of $\mathcal{N}$, the \emph{robustness measure} of ${\mathcal N}^\prime$, denoted by $\mathsf{robustness}({\mathcal N}^\prime)$, is defined to be:
\[ \mathsf{robustness}({\mathcal N}^\prime) = \frac{\sum\limits_{v_i,v_j \in V} p_{ij}({\mathcal N}^\prime[v_i,v_j])}{|{\mathcal N}^\prime|}.\]
For example, given an atomic refinement of some qualitative spatio-temporal network with a robustness measure of $0.9$, this would mean that, on average, each of its relations would have $90\%$ chance to appear in a given scenario (i.e., satisfiable atomic refinement) of that network, out of the set of all scenarios of that network. In that sense, it can be also stated that the aforementioned atomic refinement is around $90\%$ compatible with the scenario space of the network. As noted earlier, the probability $p_{ij}({\mathcal N}^\prime[v_i,v_j])$ can be approximated by ML techniques; hence, we think that this scheme can adequately support abductive learning once implemented. Clearly, based on the definition above, the most robust atomic refinement (e.g., where the probabilities are based on the output of a classifier) might not even be satisfiable. Therefore, we also argue for developing probability-based methods towards maximizing satisfiability (or, in other words, minimizing inconsistency), either via belief revision or traditional constraint propagation frameworks.

To further detail how robustness and dynamic reasoning can play a role in hybrid AI systems, let us consider the following simplified rule in the context of a train scheduling system that involves AI planning:
\begin{gather*}
  (\text{Train~$X$}~\text{\{\emph{canUse}\}}~(\text{Track~$A$}\oplus\text{Track~$B$}))\wedge(\text{Train~$Y$}~\text{\{\emph{canUse}\}}~\text{Track~$B$})~\wedge\\
  (\text{departInt(Train~$X$)}~\text{\{\emph{precedes}}_{~10\%}, \text{\emph{meets}}_{~30\%}, \text{\emph{overlaps}}_{~60\%}\}~\text{departInt(Train~$Y$)})
\end{gather*}
It is specified that Train~$X$, which uses either Track~$A$ or~$B$, has a departure slot that is $50\%$ likely (based on historical information for example) to overlap ($o$) with the departure slot of Train~$Y$, which uses Track~$B$. In the sequel, we use the linear temporal logic modalities of $\Box$ and $\Diamond$, which translate to \emph{always} and \emph{eventually} respectively. 
Naturally, some safety conditions must be met, such as the following: 
\[\Box({\neg\text{blocked(Track~$A$)}\wedge{\neg\text{blocked(Track~$B$)}}})\]
From a resource allocation perspective, it may be preferred to only use Track~$B$, and from a collision avoidance perspective, it may be preferred to use both tracks by allocating Track~$A$ to Train~$X$. In any case, a plan can be devised based on these preferences. However, let's assume that one day new information is provided by an ML system indicating that Track~$A$ will be $90\%$ likely to be subject to buckling due to a heatwave~\cite{doi:10.1177/0954409712441743}:
\[\text{Update:}~(\Diamond{\text{blocked(Track~$A$)}})_{~90\%}\]
Clearly, we now need to take this new information into account and revise our perceptions of robustness pertaining to resource allocation and collision avoidance, as well as update our original configuration. Although the example just mentioned is based on linear temporal logic (see also~\cite{DBLP:conf/frocos/BalbianiC02}), the problem applies likewise in the presence of more expressive temporal logics, such as metric temporal logic (see for example~\cite{DBLP:journals/aamas/DohertyKH09}), and when considering more complex plans. A related discussion on robust planning can be found in~\cite{hplan2021}.


\paragraph{Probabilistic infusion by use of off-the-shelf frameworks}

Extending the native QSTR framework to support probabilistic reasoning as described above, and implementing the reasoner that will support the aforementioned neuro-symbolic formulae, is a promising yet challenging task (cf.~\cite{DBLP:conf/aaaiss/GirleaA15}). As an alternative, off-the-shelf probabilistic frameworks could be leveraged, provided that spatio-temporal formalisms can be encoded using these frameworks. Luckily, a wide range of spatio-temporal formalisms, including the ones listed in Section~\ref{sec:background}, can be encoded in many different ways, most notably using Boolean satisfiability (SAT)~\cite{DBLP:conf/cp/GlorianLMS18,DBLP:conf/kr/CondottaNS16,DBLP:journals/ai/HuangLR13}, Datalog~\cite{DBLP:journals/jair/BodirskyJ17,DBLP:journals/jcss/BodirskyD13,DBLP:conf/ictai/WestphalHW13}, and Answer Set Programming (ASP)~\cite{DBLP:journals/tplp/BaryannisTBAA020,DBLP:conf/inap/JanhunenS19,DBLP:journals/tplp/WalegaSB17}. Of these frameworks, we direct the reader's attention to the latter two, since fully fledged neuro-symbolic implementation already exist, viz., DeepProbLog~\cite{DBLP:journals/ai/ManhaeveDKDR21} and NeurASP~\cite{DBLP:journals/ai/ManhaeveDKDR21}. Both of these frameworks treat the neural network output as the probability distribution over atomic facts/rules, thus providing a simple and effective way for neuro-symbolic integration. As an example, the following neuro-symbolic atom in NeurASP deals with the identification of the region variable $X$, and will apply a probability distribution over all the stated possible options:
$$
\text{nn}(\text{region}(1,X), [car, boat, train, airplane])
$$
The arbitrary options above are provided just for the sake of the example.
Likewise, a neural network that receives as input an atomic spatio-temporal network $\mathcal{N}=(V,C)$ and assigns a probability to each relation in that network (whether it holds or not) could be encoded as follows:
$$
\text{nn}(\text{network}(|C|,\mathcal{N}), [\mathsf{true}, \mathsf{false}])
$$

In relation to NeurASP, which is a probabilistic extension of ASP inspired by~\cite{DBLP:conf/kr/LeeW16}, exploring other possibilities with probabilistic ASP, e.g., \cite{DBLP:conf/cilc/AzzoliniBR22}, could be a fruitful research direction.
Similar examples can be formed in DeepProbLog, subject to different syntax. Arguably, a systematic analysis of the effect of the spatio-temporal relations/constraints might be needed to make these frameworks scale in our context, but we think they could serve as a good starting point for neuro-symbolic spatio-temporal reasoning via probabilities. 


\subsection{Datasets}
\label{sec:datasets}


\begin{figure}[t]
  \centering%
  \subfloat[%
  \textbf{Q:} ``What size is the cylinder that is left of the brown metal thing that is left of the big sphere?'' \textbf{A:} ``Small.''\label{fig:clevr}%
  ]{%
    \centering%
    \includegraphics[width=0.3\linewidth]{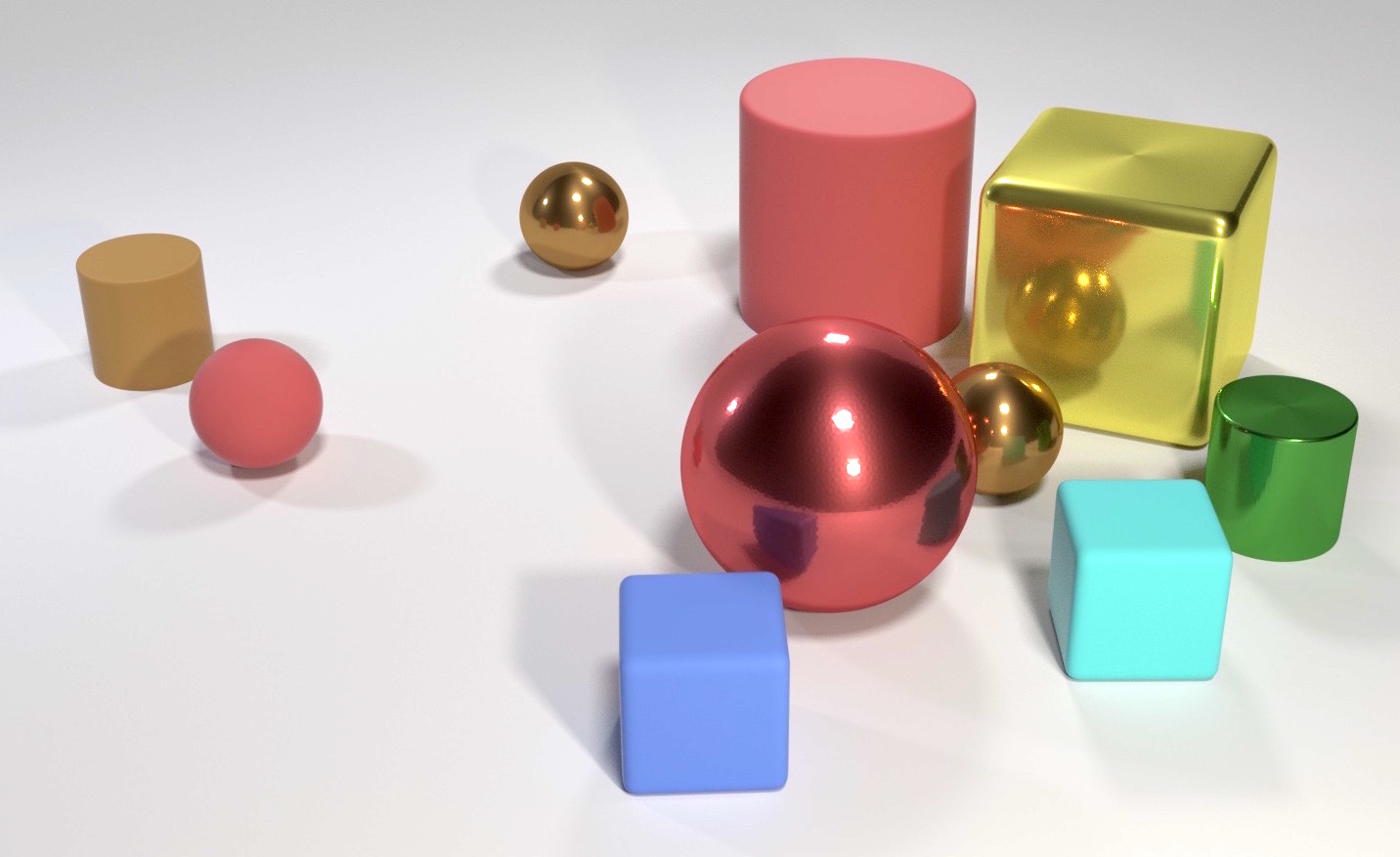}%
  }%
  \hfill%
  \subfloat[%
  \textbf{Q:} ``Clock aligned to Camera?'' \textbf{A:} ``Yes.''\label{fig:rel3d}%
  ]{%
    \centering%
    \includegraphics[width=0.3\linewidth]{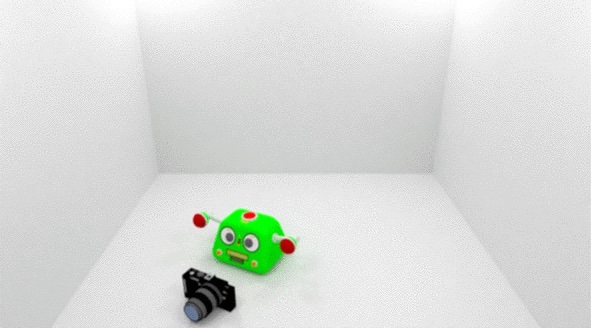}%
  }%
  \hfill%
  \subfloat[%
  \textbf{Q:} ``How many objects with purple legs are there?'' \textbf{A:} ``2'' \label{fig:ptr}%
  ]{%
    \centering%
    \includegraphics[width=0.3\linewidth]{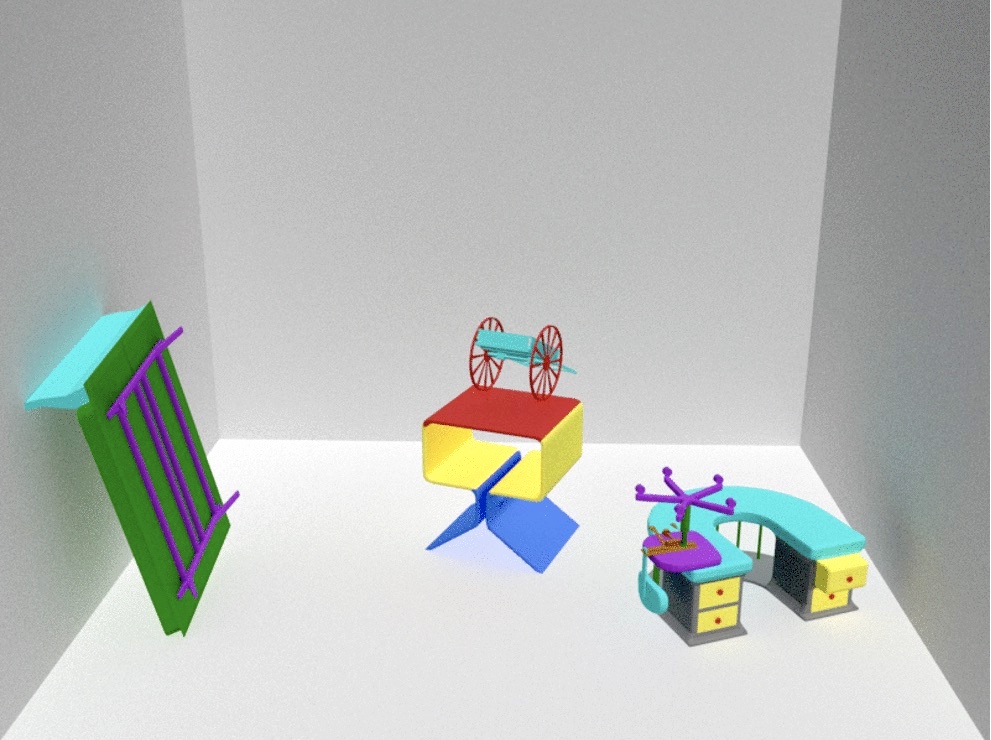}%
  }%
  \caption{Some examples from datasets CLEVR~\cite{johnson_clevr_2017}, Rel3D~\cite{goyal_rel3d_2020} and PTR~\cite{hong_ptr_2021}.}
  \label{fig:datasets}
\end{figure}

Spatial relations are at the core of spatio-temporal reasoning. Recognizing spatial relations is the first step towards solving problems that involve physical objects, such as \emph{visual question answering} (VQA), which is concerned with answering questions about an image~\cite{wu_visual_2017}. Several datasets have been proposed for VQA~(e.g., \cite{goyal_making_2017,johnson_clevr_2017,suhr_corpus_2019,hudson_gqa_2019}), which are either based on real-world images and crowdsourced questions\ms{,} or synthetic images with automatically generated questions that often require complex reasoning. The task \ms{pertaining to} the latter datasets \ms{is} also known as \emph{visual reasoning}. As neuro-symbolic approaches are amenable to the visual reasoning task, several neuro-symbolic approaches have been developed for visual reasoning~\cite{yi_neural-symbolic_2018,mao_neuro-symbolic_2018,amizadeh_neuro-symbolic_2020}.

A task that is specialized in relation learning is \emph{visual relationship detection}~(VRD)~\cite{lu_visual_2016,agarwal_visual_2020,krishna_visual_2017}, where the goal is to predict the relation between objects in an image. Unlike VQA, the inputs to VRD are just images, thus neglecting the expressive power of predicates describing relative spatial relations framed within a rich linguistic context. The outputs also differ from those of VQA, as they are the bounding boxes of the objects in the image, their labels\ms{,} and the relations between the objects.

VQA datasets, such as CLEVR (cf.~Figure~\ref{fig:clevr}), typically use spatial relations in an absolute frame of reference (i.e., \emph{left} and \emph{right} correspond to the \emph{left} and \emph{right} side of the input image) and do not deal with other types of relations, such as part-whole relations. Moreover, datasets using crowdsourced questions introduce specific biases that can be exploited by the models.

Spatial relations in visual question answering are often easily predicted using simple cues (e.g., 2D spatial configurations or language priors). To mitigate these issues, the SpatialSense~\cite{yang_spatialsense_2019} dataset was created, where human annotators provide spatial relation labels that are difficult to predict using these cues. Rel3D~\cite{goyal_rel3d_2020} (cf.~Figure~\ref{fig:rel3d}) is also a crowdsourced dataset, but unlike SpatialSense, the images in Rel3D are synthetically generated using 3D ground-truth information and provide reduced annotation bias. The dataset is limited in that the images in \ms{it} contain only two objects\ms{,} and only one type of question is asked, namely predicting the correctness of triples \ms{of the form} (object\textsubscript{1}, relation, object\textsubscript{2}).

The GRiD-3D dataset that is introduced in Section~\ref{sec:symb-meta-learn} provides a different number of objects in each scene and different question categories. The variety of question categories offered by GRiD-3D helps neural models learn directional relationships in a multi-task environment without the need for additional 3D ground-truth information.

The PTR dataset~\cite{hong_ptr_2021}~(cf.~Figure~\ref{fig:ptr}) focuses on the part-whole relationships between entities in synthetic scenes. A question in the PTR dataset is for example ``How many objects with purple legs are there?''. The CLEVR Mental Rotation Tests~\cite{christopher_beckham_visual_2021} investigate the role of the viewer's perspective in visual question answering.



\section{Conclusion}
\label{sec:conclusion}

In this chapter, we discussed the problem of interrelating machine learning with spatial and temporal reasoning from the broader perspective of Neuro-Symbolic AI, which is an area of AI that focuses on the seamless integration of statistical learning and symbolic reasoning. It is worth noting that this integration problem of spatial and temporal reasoning and learning forms an active and appealing research topic to the AI community and provides many opportunities for cross-disciplinary collaboration and research~\cite{DBLP:conf/ijcai/2022strl,DBLP:conf/ijcai/2018lr}.

Our main rationale behind devoting a research work that is particular to spatial and temporal information 
is that the challenging real-world problems of today, such as obtaining self-driving vehicles, comprise several spatio-temporal aspects, e.g., topology, orientation, and time order, to name a few, whose study has its own large share in the literature~\cite{dylla_survey_2017,ligozat2011qualitative,cohn_qualitative_2008,DBLP:reference/spatial/2007}. This is because spatial and temporal reasoning is in and of itself a large research domain in AI that involves several logic properties, theories, and characteristics strictly pertaining to the infinite domains of space and time. Therefore, in order to have an efficient and effective synergy between logical reasoning and machine learning that will be grounded on spatial and temporal knowledge, it is important to leverage the representation and reasoning techniques that have been made available over the past decades from the spatial and temporal reasoning community. In addition, we argued that spatial and temporal knowledge encodes hard and verifiable knowledge that an AI agent should inherently possess, or be able to learn through interacting with the physical environment, and thus it forms an integral part of the generalization capabilities of that agent. Indeed, spatio-temporal formalisms encode valid assumptions that align with our understanding of physics and human cognition, and a learner that does not use them should fare worse than one that does.

To this end, we proposed a synergy between logical reasoning and machine learning that is grounded on spatial and temporal knowledge and, by extension, on physics and human cognition. Specifically, we provided a brief overview of neuro-symbolic AI and the rich area of qualitative spatio-temporal reasoning, and we introduced and formalized some works on neuro-symbolic spatio-temporal reasoning and learning. \ms{In particular, we looked into a semantic referee for geospatial semantic segmentation of satellite imagery data, and a symbolic meta-learning interface for learning relative directions. Further, we proposed a generic neuro-symbolic framework for integrating qualitative spatio-temporal reasoning and neural methods from a probabilistic perspective to deal with uncertainty. Finally, we also reviewed recent datasets for spatial relation learning.}

\jl{A potential future direction could be devising} a complete probabilistic framework for handling uncertainty in spatio-temporal reasoning and learning, as described in~\ref{sec:prob}, and apply it \jl{to real-world domains}, such as the domain of embodied robotics~\cite{lee_spatial_2022} or self-driving vehicles, which involves many problems of that nature. Consider the case of tree shadows as an example, which are known to cause unnecessary or overly cautious braking in self-driving vehicles that could pose crash risks in heavy traffic~\cite{iihs2018,michael_j_coren_all_2018}. 
This problem is due to the fact that the ML techniques in the vehicles often confuse shadows with some kind of obstacle (cf.~\cite{DBLP:journals/semweb/AlirezaieLSL19} where shadows are classified as water). Therefore, it is important to provide \jl{a framework to integrate} such techniques with spatial and temporal reasoning, which should rectify the output of the classifier at runtime. As another example, we can recall the case of a Tesla Model~$3$ car that was detecting an endless trail of traffic lights while it was driving behind a truck hauling deactivated traffic lights~\cite{dan_robitzski_watch_2021}. 
Again, such edge cases, that are in fact hidden vulnerabilities of an ML model, will always be a cause of concern in end-to-end learning systems, as sufficient data diversity is an untestable assumption a priori~\cite{DBLP:journals/pieee/ScholkopfLBKKGB21}.

\jl{As recent research in representation learning has shown, pretraining on a \emph{large set of data} is often the key to better performance and robustness; real-world data is however difficult or expensive to obtain. By contrast, synthetic datasets are cheaper to obtain and allow for incorporating the underlying causal structures that constitute a spatial scene (e.g., object position/orientation/color/shape, camera perspective) which can be formulated symbolically. Therefore, we believe that generating synthetic data based on symbolic causal considerations~\cite{kaddour_causal_2022,DBLP:journals/pieee/ScholkopfLBKKGB21} (e.g., to provide an implicit curriculum like in Section~\ref{sec:symb-meta-learn}) can be another promising future direction.}


\bibliographystyle{tfnlm} %
\bibliography{chapter-dummy/jae,chapter-dummy/michael}


\end{document}